\documentclass[10pt,twocolumn,letterpaper]{article}

\usepackage{iccv}
\usepackage{times}
\usepackage{epsfig}
\usepackage{graphicx}
\usepackage{amsmath}
\usepackage{amssymb}
\usepackage{bbm}
\usepackage{epstopdf}
\usepackage{caption}
\usepackage{subcaption}
\usepackage{enumitem}
\usepackage{calc}
\usepackage{multirow}
\usepackage{xspace}
\usepackage{booktabs}
\usepackage{mathrsfs}
\usepackage{array}

\usepackage[breaklinks=true,bookmarks=false]{hyperref}

\iccvfinalcopy 

\def\iccvPaperID{1581} 

\ificcvfinal\fi

\begin{document}

\title{Towards Precise End-to-end Weakly Supervised Object Detection
	Network}

\author{Ke Yang \qquad Dongsheng Li \qquad Yong Dou\\
National University of Defense Technology\\
{\tt\small yangke13@nudt.edu.cn}}

\maketitle
\ificcvfinal\thispagestyle{empty}\fi

\begin{abstract}
It is challenging for weakly supervised object detection network to precisely predict the positions of the objects, since there are no instance-level category annotations. Most existing methods tend to solve this problem by using a two-phase learning procedure, i.e., multiple instance learning detector followed by a fully supervised learning detector with bounding-box regression. Based on our observation, this procedure may lead to local minima for some object categories. In this paper, we propose to jointly train the two phases in an end-to-end manner to 	tackle this problem. Specifically, we design a single network with both multiple instance learning and bounding-box regression branches that share the same backbone. Meanwhile, a guided attention module using classification loss is added to the backbone for effectively extracting the implicit location information in the features. Experimental results on public datasets show that our method achieves state-of-the-art performance.
\end{abstract}
\section{Introduction}
In recent years, Convolutional Neural Networks (CNN) approaches have achieved great success in computer vision field, due to its ability to learn generic visual features that can be applied in many tasks such as image classification \cite{alexnet,vgg,resnet}, object detection \cite{rcnn,fast-rcnn,faster-rcnn} and semantic segmentation \cite{fcn,deeplabv1}. Fully supervised object detection has been widely studied and achieved promising results. There are also plenty of public datasets which provide precise location and category annotations of the objects. However, precise object-level annotations are always expensive in 
human resource and huge data volume is required by training accurate object detection models. In this paper, we focus on Weakly Supervised Object Detection (WSOD) problem, which uses only image-level category labels so that significant cost of preparing training data can be saved. Due to the lack of accurate annotations, this problem has not been well handled and the performance is still far from the fully supervised methods.

\begin{figure}[t]
\centering
	\includegraphics[width = 0.45\textwidth]{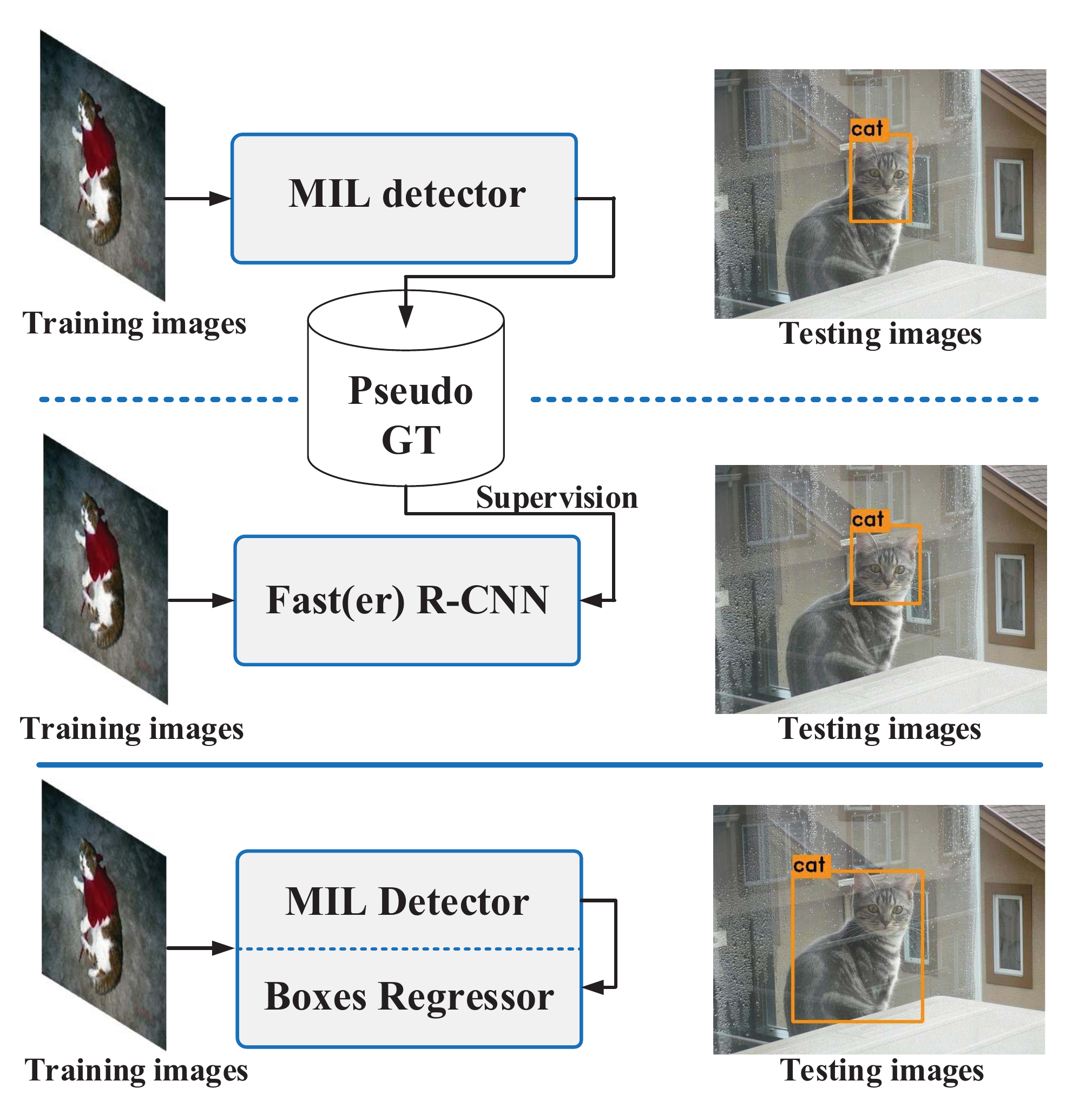}
	\caption{The learning strategy comparison of existing weakly supervised
		object detection methods (above the blue solid line) and our proposed method (below the blue solid line).}
	\label{fig1}
\end{figure}
\begin{figure*}[t]
    \centering
	\includegraphics[width = 1.0\textwidth]{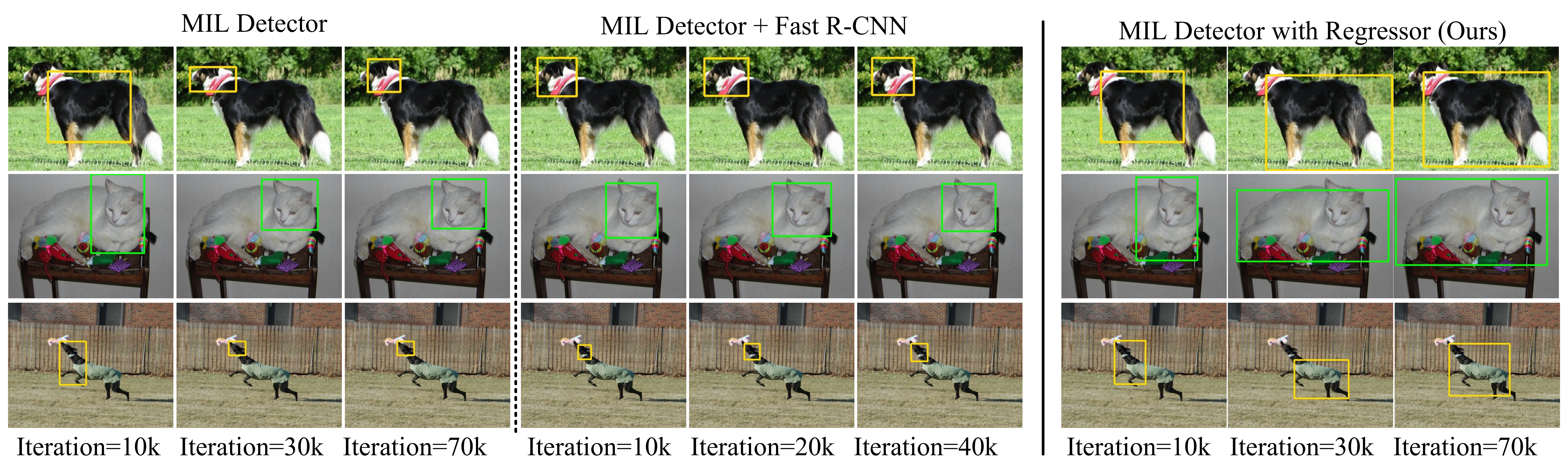}
	\caption{Detection results of MIL detector (left part), Fast R-CNN with pseudo GT from MIL detector (middle part) and our jointly training network (right part)  at different training iterations.}
	\label{fig2}
\end{figure*}
Recent WSOD methods \cite{wccn,wsddn,oicr,pda,selftaught} usually follows a
two-phase learning procedure as shown in the top part of Figure \ref{fig1}. In the first phase, the Multiple Instance Learning (MIL) \cite{mfmi,selftaught,oicr,wsddn} like weakly learning pipeline is used, which trains a MIL detector by using CNN as feature extractor. In the second phase, a fully supervised detector, e.g. Fast R-CNN \cite{fast-rcnn} or Faster R-CNN \cite{faster-rcnn}, is trained to further refine object location by using the selected proposals of the first phase as supervision. The main functionality of the second phase is to regress the object locations more precisely. However, we observed that the two-phase learning is easy to get stuck into local minima if the selected proposals of the first phase are too far from real Ground Truth (GT). As shown in the top part of Figure \ref{fig1}, in some categories, the MIL detector tends to focus on the local discriminative parts of the objects, such as the head of a cat, so that the wrong proposals are used as pseudo GT for the second phase. In this case, the accurate location of the object can hardly be learned in the regression process of the second phase, as the MIL detector has already over-fitted seriously to the discriminate parts, as shown in the middle part of Figure \ref{fig2}.

We further observed that the MIL detector does not select the most discriminative parts
at the beginning of the training, but gradually over-fits to these parts, as shown in the left part of Figure \ref{fig2}.

Taking into account the above observations, we propose to jointly train the MIL detector and the bounding-box regressor together in an end-to-end manner, as shown in the bottom part of Figure \ref{fig1}. In this manner, the regressor is able to start to adjust the predicted boxes before the MIL detector focuses seriously to small discriminative parts, as shown in the right part of Figure \ref{fig2}. 
Specifically, we use MIL detection scheme \cite{wsddn,oicr} as baseline and integrate fully supervised RoI-based classification and bounding-box regression branch similar to Fast R-CNN, which shares the same backbone with MIL detector. MIL detector is a weakly learning process, which selects object predictions from the region proposals, e.g. generated by Selective Search Windows (SSW) \cite{ss} method, according to classification scores. These selected proposals are then used as the pseudo GT supervision of the classification and regression branch.

In order to further enhance the localization ability of the proposed network, we propose to use a guided attention module using image-level classification loss in the backbone. To our best knowledge, the well trained classification network contains rich object location information. Therefore, we add this attention branch which is guided by image-level classification loss. Fully considering the global characteristics of the objects, the attention branch can improve the discriminative ability of the network as well as detection accuracy.

It is worth noting that though jointly learning of classification and boxes regression has already been shown to be beneficial for fully supervised object detection, for weakly supervised object detection it is still non-trivial and needs innovative idea and insight on this task. Although Our method is conceptually simple in form, it significantly alleviates the weak detector over-fitting to discriminate parts and substantially surpasses previous methods.
Our contributions can be summarized as follows.
\begin{itemize}
	\item We design a single end-to-end weakly supervised object detection network that can jointly optimize the region classification and regression, which boosts performance significantly.
	\item We design a classification guided attention module to enhance the localization ability of feature learning, which also leads to a noteworthy improvement.
	\item Our proposed network significantly outperforms previous state-of-the-art weakly supervised object detection approaches on PASCAL VOC 2007 and 2012.
\end{itemize} 
\section{Related Work}
\subsection{Convolutional Feature Extraction}
After the success of using CNNs for image classification task\cite{alexnet}, a research stream based on CNNs \cite{rcnn,overfeat} shows significant improvements in detection performance. These methods use convolutional layers to extract features from each region proposal. To speed up the the detection, SPP-Net~\cite{sppnet} and Fast R-CNN~\cite{fast-rcnn} firstly extract region-independent feature maps at the full-image level, and then pool region-wise features via spatial extents of proposals.
\subsection{Weakly Supervised Object Detection}
Most existing methods formulate weakly-supervised detection as a multiple instance learning problem \cite{wsddn,song2014weakly,hoffman2015detector,selftaught,pda,appearancetransfer}. These approaches divided training images into positive and negative parts, where each image is considered as a bag of candidate object instances. If an image is annotated as a positive sample of a specific object class, at least one proposal instance of the image belongs to this class. The main task of MIL-based detectors is to learn the discriminative representation of the object instances and then select them from positive images to train a detector. Previous works on applying MIL to WSOD can be roughly categorized into \textbf{multi-phase learning approach} \cite{selftaught,mfmi,pda,min-entropy,gal-fwsd,ml-locanet,w2f,zigzag} and \textbf{end-to-end learning approach} \cite{wsddn,ts2c,oicr,contextlocnet,pcl}.

\noindent \textbf{End-to-end learning approaches} combine CNNs and \textbf{MIL} into a unified network to address weakly supervised object detection task. Diba \textit{et al.} \cite{wccn} proposed an end-to-end cascaded convolutional network to perform weakly supervised object detection and segmentation in cascaded manner. Bilen \textit{et al.} \cite{wsddn} developed a two-stream weakly supervised deep detection network (WSDDN), which selected the positive samples by aggregating the score of classification stream and detection stream. Based on WSDDN, Kantorov \textit{et al.} \cite{contextlocnet} proposed to learn a context-aware CNN with contrast-based contextual modeling. Also based on WSDDN, Tang \textit{et al.} \cite{oicr} designed an online instance classifier refinement (OICR) algorithm to alleviate the local optimum problem. Tang \textit{et al.} \cite{pcl} also proposed Proposal Cluster Learning (PCL) to improve the performance of OICR. Following the inspiration of \cite{contextlocnet} and \cite{wccn}, Wei \textit{et al.} \cite{ts2c} proposed a tight box mining method that leverages surrounding segmentation context derived from weakly-supervised segmentation to suppress low quality distracting candidates and boost the high-quality ones. Recently, Tang \textit{et al.} \cite{wsrpn} proposed a weakly supervised region proposal network to generate more precise proposals for detection. Positive object instances often focus on the most discriminative parts of an object (e.g. the head of a cat, etc.) but not the whole object, which leads to inferior performance of weakly supervised detectors.

\noindent \textbf{Multi-phase learning approaches} first employ MIL to select the best object candidate proposals, then use these selected proposals as pseudo GT annotations for learning the fully supervised object detector such as R-CNN \cite{rcnn} or Fast(er) R-CNN \cite{fast-rcnn,faster-rcnn}. Li \textit{et al.} \cite{pda} proposed classification adaptation to fine-tune the network to collect class specific object proposals, and detection adaptation was used to optimize the representations for the target domain by the confident object candidates. Cinbis \textit{et al.} \cite{mfmi} proposed a multi-fold MIL detector by re-labeling proposals and re-training the object classifier iteratively to prevent the detector from being locked into wrong object locations. Jie \textit{et al.} \cite{selftaught} proposed a self-taught learning approach to progressively harvest high-quality positive instances. Zhang \textit{et al.} \cite{w2f} proposed pseudo ground-truth excavation (PGE) algorithm and pseudo groundtruth adaptation (PGA) algorithm to refine the pseudo ground-truth obtained by \cite{oicr}. Wan \textit{et al.} \cite{min-entropy} proposed a min-entropy latent model (MELM) and recurrent learning algorithm for weakly supervised object detection. Ge \textit{et al.} \cite{multi-evidence} proposed to fuse and filter object instances from different techniques and perform pixel labeling with uncertainty and they used the resulting pixelwise labels to generate groundtruth bounding boxes for object detection and attention maps for multi-label classification. Zhang \textit{et al.} \cite{ml-locanet} proposed a Multi-view Learning Localization Network (ML-LocNet) by incorporating multiview learning into a two-phase WSOD model. However, multi-phase learning WSOD is a non-convex optimization problem, which makes such approaches trapped in local optima.

In this paper, we consider the MIL (positive object candidates mining) and regression (object candidates localization refinement) problems simultaneously. We follow the MIL pipeline and combine the two-stream WSDDN \cite{wsddn} and OICR/PCL algorithms \cite{oicr,pcl} to implement our basic MIL branch and refine the detected boxes with a regression branch in an online manner.
\subsection{Attention Module}

Attention modules were first used in the natural language processing field
and then introduced to the computer vision area. Attention can be seen as a method of biasing the allocation of available computational resources towards the most informative components of a signal
\cite{att-history1,att-history2,att-history3,att-history4,self-att,recurrent-att,senet}.

The current attention modules can be divided into two categories: spatial attention and channel-wise attention. Spatial attention is to assign different weights to different spatial regions depending on their feature content. It automatically predicts the weighted heat map to enhance the relevant features and suppress the irrelevant features during the training process of a specific task. Spatial attention has been used in image captioning \cite{xu2015show}, multi-label classification \cite{zhu2017learning}, pose estimation \cite{chu2017multi} and so on. Hu \textit{et al.} \cite{senet} proposed an Squeeze-and-Excitation block which models channel-wise attention in a computationally efficient manner. In this paper, we use a combination of spatial and channel-wise attention, and \textit{our attention module is guided by object category}. 
\begin{figure*}[t]
\centering
\includegraphics[width = 1.0\textwidth]{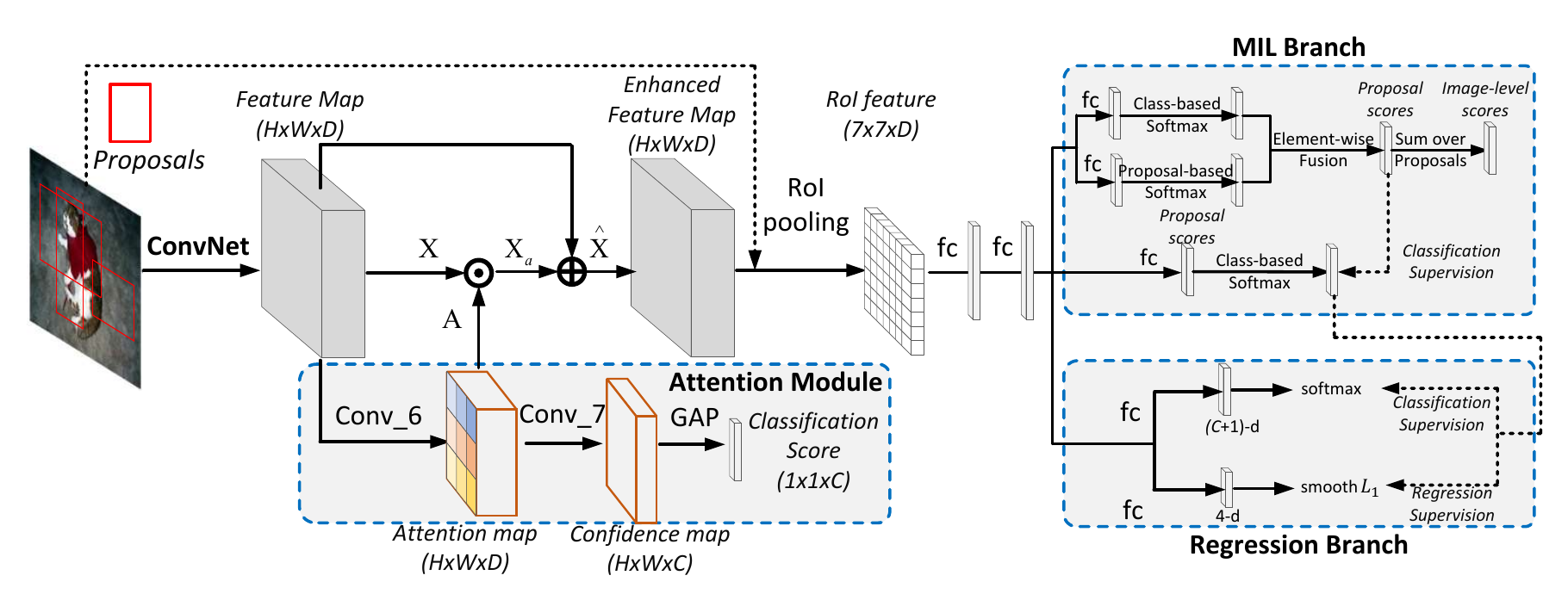}
\caption{Architecture of our proposed network. (1) Generate
	discriminate
	features using attention mechanism. (2) Generate the RoI
	features from enhanced feature map. (3) {\bf MIL branch}: Feed the
	extracted RoI features
	into a MIL network for pseudo GT boxes annotation initialization.
	(4)
	{\bf Regression branch}: Feed the extracted RoI features and
	generated pseudo GT to the regression branch for RoI classification
	and regression. }
\label{fig:overall-arch}
\end{figure*}
\section{Method}
In this section we introduce proposed weakly supervised object
detection network, which consists of three major components: guided
attention module (GAM), MIL branch and regression branch. The overall
architecture of proposed network is shown in Figure \ref{fig:overall-arch}. Given an input image, an enhanced feature map is first extracted from the CNN network with GAM. Region features generated by ROI pooling are then sent to MIL branch and regression branch. The object locations and categories proposed by MIL branch are taken as pseudo GT of the regression branch for location regression and classification. The remainder of this section discusses the three components in detail.
\subsection{Guided Attention Module}\label{sec:gam}
First, we describe the conventional spatial neural attention structure.
Given a feature map $\textbf{X} \in \mathbb{R}^{H \times W \times D}$ extracted from a ConvNet, the attention module takes it as input and outputs a spatial-normalized attention weight map $\textbf{A} \in \mathbb{R}^{H \times W}$ via a $1$$\times$$1$ convolutional layer. Attention map is then multiplied to $\textbf{X}$ to get attended feature  $\textbf{X}_a \in \mathbb{R}^{H \times W \times D}$. $\textbf{X}_a$ is added to $\textbf{X}$ to get the enhanced feature map $\hat{\textbf{X}}$. After that, $\hat{\textbf{X}}$ is fed to subsequent modules. Attention map $\textbf{A}$ acts as a spatial regularizer to enhance the relevant regions and suppress the non-relevant regions for feature $\textbf{X}$.

Formally, attention module consists of a convolutional layer, a non-linear
activation layer and a spatial normalization as follows:
\begin{equation}\label{eq:conv_sigmoid}
z_{i,j} = F\left(\textbf{w}^\textbf{T} {\textbf{x}}_{i,j} + b\right),
\end{equation}
\begin{equation}\label{eq:att}
a_{i,j}= \frac{z_{i,j}}{\sum_{i,j} z_{i,j}},
\end{equation}
\noindent where $F$ is non-linear activation function. $\textbf{w}$ and $b$ are the parameters of the attention module, which is a $1\times1$ convolutional layer. The attended feature $\hat{\textbf{x}}_{i,j}$ can
be calculated by:
\begin{equation}\label{eq:att_muladd}
{\hat{\textbf{x}}}_{i,j}= (1+{a}_{i,j}){\textbf{x}}_{i,j}.
\end{equation}

The conventional attention map is class-agnostic. We hope it can learn some
foreground/background information to help figure out the position of the objects, because it has been proved that CNNs are not only effective at predicting the class label of an image, but also localizing the image regions relevant to this label \cite{cam}.

We add the classification loss to guide the learning of the attention weights. To achieve this, we expand spatial attention to both spatial and
channel attention. Specifically, attention map are changed from $\textbf{A} \in \mathbb{R}^{H \times W}$ to  $\textbf{A}\in
\mathbb{R}^{H \times W \times D}$. The attention module can be formalized
as:
\begin{equation}\label{eq:conv_sigmoid_2}
{z}^c_{i,j} = F\left(\textbf{w}_c^\textbf{T} {\textbf{x}}_{i,j} + \textbf{b}^c\right),
\end{equation}

\begin{equation}\label{eq:att_2}
a^c_{i,j}= \frac{{z}^c_{i,j}}{1 + \exp{(-{z}^c_{i,j})}},
\end{equation}
where $c$ denotes the value of the $c$-th channel. The attended feature
$\hat{\textbf{x}}^c_{i,j}$ can be calculated by:
\begin{equation}\label{eq:att_muladd_2}
\hat{\textbf{x}}^c_{i,j} = (1+{a}_{i,j}^c){\textbf{x}}_{i,j}^c.
\end{equation}

To introduce classification supervision to attention weights learning, attention map $\textbf{A}$ is also fed to another convolutional layer and a Global Average Pooling (GAP) layer to get the classification score vector. Then the attention map can be supervised by the standard multi-label classification loss. The enhanced feature map $\hat{\textbf{X}}$ is fed to subsequent components for detection.
\begin{table*}[]
\centering
\fontsize{7}{8}\selectfont
\def\arraystretch{1.1}
\setlength{\tabcolsep}{3pt}
\begin{tabular}{l|cccccccccccccccccccc|c}
\specialrule{.2em}{.1em}{.1em}
Methods&aero&bike&bird&boat&bottle&bus&car&cat&chair&cow&table&dog&horse&mbike&person&plant&sheep&sofa&train&tv&mAP
\\\hline
MIL&56.2&62.1&39.4&21.8&10.3&63.6&60.6&31.8&24.8&45.9&35.3&24.1&36.7&63.3&13.1&23.1&39.4&49.1&64.7&60.3&41.3\\
MIL+GAM&55.2&62.5&42.6&23.0&12.7&66.2&62.0&39.2&26.1&48.9&37.7&26.1&45.3&64.5&12.8&24.4&42.3&46.4&65.9&62.4&43.3\\
MIL+FRCN&60.2&65.0&50.9&24.9&11.9&71.6&68.0&34.6&27.2&61.2&40.8&17.6&47.1&65.6&13.0&22.8&51.0&57.6&66.5&60.5&45.9\\
MIL+REG&56.5&63.4&38.8&28.3&15.3&68.2&66.6&68.0&23.7&51.6&46.0&32.4&53.8&63.9&12.1&23.5&47.2&56.3&65.2&64.9&47.3\\
MIL+GAM+REG&55.2&66.5&40.1&31.1&16.9&69.8&64.3&67.8&27.8&52.9&47.0&33.0&60.8&64.4&13.8&26.0&44.0&55.7&68.9&65.5&48.6\\
\hline
\end{tabular}
\caption{Ablation study: AP performance (\%) on PASCAL VOC 2007 test}
\label{table:ablation_study}
\end{table*}
\begin{table*}[]
\centering
\fontsize{7}{8}\selectfont
\def\arraystretch{1.1}
\setlength{\tabcolsep}{3pt}
\begin{tabular}{l|cccccccccccccccccccc|c}
\specialrule{.2em}{.1em}{.1em}
Methods&aero&bike&bird&boat&bottle&bus&car&cat&chair&cow&table&dog&horse&mbike&person&plant&sheep&sofa&train&tv&mean
\\\hline
MIL&82.5&76.5&61.0&47.3&30.2&80.7&82.4&44.8&42.1&78.1&45.2&32.8&57.1&89.6&16.6&57.9&73.2&61.8&79.1&73.5&60.6\\
MIL+GAM&82.1&78.4&64.3&48.9&32.4&81.2&82.9&48.5&43.4&79.5&43.7&34.9&61.9&89.2&16.6&57.5&71.1&56.2&78.7&77.4&61.4\\
MIL+FRCN&83.8&81.2&65.2&48.4&34.4&84.3&84.6&49.4&44.8&82.9&48.7&37.7&67.0&90.0&21.4&60.1&76.3&66.4&82.5&80.6&64.5\\
MIL+REG&82.1&79.2&61.6&52.7&33.2&82.7&85.8&77.3&39.2&82.2&47.5&42.3&75.2&92.0&19.3&58.6&79.4&65.6&77.2&83.9&65.8\\
MIL+GAM+REG&81.7&81.2&58.9&54.3&37.8&83.2&86.2&77.0&42.1&83.6&51.3&44.9&78.2&90.8&20.5&56.8&74.2&66.1&81.0&86.0&66.8\\
\hline
\end{tabular}
\caption{Ablation study: CorLoc performance (\%) on PASCAL VOC 2007 trainval}
\label{table:ablation_study_corloc}
\end{table*}
\subsection{MIL Branch}\label{sec:pgm}
We only have image-level labels indicating whether an object category appears. To train a standard object detector with regression, it is necessary to mine instance-level supervision such as bounding-box annotations. Therefore, we need to introduce a MIL branch to initialize the pseudo GT annotations. There are a couple of possible choices such as \cite{wsddn,mfmi,oicr}. We choose to adopt OICR network \cite{oicr} which is based on WSDDN \cite{wsddn} for its effectiveness and end-to-end training. WSDNN employed a two streams network: the classification and detection data streams. By aggregating these two streams, instance-level predictions can be achieved.

Specifically, given an image $\textbf{I}$ with only image-level label $\textbf{Y}=[y_1, y_2,...,y_C] \in \mathbb{R}^{C\times1}$, where $y_c=1$ or $0$ indicates the presence or absence of an object class $c$. For each input image $\textbf{I}$, the object proposals $\mathcal{R} = (R_1, R_2,...,R_n)$ are generated by the selective search windows method \cite{ss}. The features of each proposal are extracted through a ConvNet pre-trained on ImageNet \cite{imagenet} and RoI Pooling, then are branched into two streams to produce two matrices $\textbf{x}^{cls},\textbf{x}^{det} \in \mathbb{R}^{C \times |\mathcal{R}|}$ by two FC layers, where $|\mathcal{R}|$ denotes the number of proposals and $C$ denotes the number of image classes. These two matrices are passed through a softmax layer with different dimensions and the outputs are two matrices with the same shape: $\sigma(\textbf{x}^{det})$ and $\sigma(\textbf{x}^{cls})$.

After that, the scores of all proposals are generated by element-wise product $\textbf{x}^\mathcal{R} = \sigma(\textbf{x}^{det}) \odot \sigma(\textbf{x}^{cls})$. Finally, the $c$-th class prediction score at the image-level can be obtained by summing up the scores over all proposals: $p_c=\sum_{r=1}^{|\mathcal{R}|} \textbf{x}^\mathcal{R}_{c,r}$ .

During the training stage, the loss function can be formulated as follows:
\begin{equation}\label{eq:loss_w}
\mathcal{L}_{mil} = - \sum_{c=1}^{C}\{y_c\log p_c \ + \
(1-y_c)\log(1-p_c)\}.
\end{equation}

Since the performance of WSDDN is unsatisfactory, we adopt the OICR \cite{oicr} and its upgraded version Proposal Cluster Learning (PCL) \cite{pcl} to refine the proposal classification results of WSDDN.

After several times classifier refinement, the classifier tends to select the tight boxes as positive instances, which can be used as pseudo GT annotations for our online boxes regressor.
\begin{table*}[]
\centering
\fontsize{7}{8}\selectfont
\def\arraystretch{1.1}
\setlength{\tabcolsep}{2.5pt}
\begin{tabular}{l|cccccccccccccccccccc|c}
\specialrule{.2em}{.1em}{.1em}
Methods&aero&bike&bird&boat&bottle&bus&car&cat&chair&cow&table&dog&horse&mbike&person&plant&sheep&sofa&train&tv&mAP
\\\hline
WSDDN\cite{wsddn}&39.4&50.1&31.5&16.3&12.6&64.5&42.8&42.6&10.1&35.7&24.9&38.2&34.4&55.6&9.4&14.7&30.2&40.7&54.7&46.9&34.8\\
ContextLocNet\cite{contextlocnet}&57.1&52.0&31.5&7.6&11.5&55.0&53.1&34.1&1.7&33.1&49.2&42.0&47.3&56.6&15.3&12.8&24.8&48.9&44.4&47.8&36.3\\
OICR\cite{oicr}&58.0&62.4&31.1&19.4&13.0&65.1&62.2&28.4&24.8&44.7&30.6&25.3&37.8&65.5&15.7&24.1&41.7&46.9&64.3&62.6&41.2\\
Self-taught\cite{selftaught}&52.2&47.1&35.0&26.7&15.4&61.3&66.0&54.3&3.0&53.6&24.7&43.6&48.4&65.8&6.6&18.8&51.9&43.6&53.6&62.4&41.7\\
WCCN\cite{wccn}&49.5&60.6&38.6&29.2&16.2&70.8&56.9&42.5&10.9&44.1&29.9&42.2&47.9&64.1&13.8&23.5&45.9&54.1&60.8&54.5&42.8\\
TS2C\cite{ts2c}&59.3&57.5&43.7&27.3&13.5&63.9&61.7&59.9&24.1&46.9&36.7&45.6&39.9&62.6&10.3&23.6&41.7&52.4&58.7&56.6&44.3\\
WSRPN\cite{wsrpn}&57.9&70.5&37.8&5.7&21.0&66.1&69.2&59.4&3.4&57.1&57.3&35.2&64.2&68.6&32.8&28.6&50.8&49.5&41.1&30.0&45.3\\
PCL\cite{pcl}&54.4&69.0&39.3&19.2&15.7&62.9&64.4&30.0&25.1&52.5&44.4&19.6&39.3&67.7&17.8&22.9&46.6&57.5&58.6&63.0&43.5\\\hline
MIL-OICR+GAM+REG(Ours)&55.2&66.5&40.1&31.1&16.9&69.8&64.3&67.8&27.8&52.9&47.0&33.0&60.8&64.4&13.8&26.0&44.0&55.7&68.9&65.5&48.6\\
MIL-PCL+GAM+REG(Ours)&57.6&70.8&50.7&28.3&27.2&72.5&69.1&65.0&26.9&64.5&47.4&47.7&53.5&66.9&13.7&29.3&56.0&54.9&63.4&65.2&51.5\\\hline\hline
PDA\cite{pda}&54.5&47.4&41.3&20.8&17.7&51.9&63.5&46.1&21.8&57.1&22.1&34.4&50.5&61.8&16.2&29.9&40.7&15.9&55.3&40.2&39.5\\
WSDDN-Ens.\cite{wsddn}&46.4&58.3&35.5&25.9&14.0&66.7&53.0&39.2&8.9&41.8&26.6&38.6&44.7&59.0&10.8&17.3&40.7&49.6&56.9&50.8&39.3\\
OICR-Ens.+FRCNN\cite{oicr}&65.5&67.2&47.2&21.6&22.1&68.0&68.5&35.9&5.7&63.1&49.5&30.3&64.7&66.1&13.0&25.6&50.0&57.1&60.2&59.0&47.0\\
WCCN+FRCNN\cite{wccn}&-&-&-&-&-&-&-&-&-&-&-&-&-&-&-&-&-&-&-&-&43.1\\
MELM\cite{multi-evidence}&55.6&66.9&34.2&29.1&16.4&68.8&68.1&43.0&25.0&65.6&45.3&53.2&49.6&68.6&2.0&25.4&52.5&56.8&62.1&57.1&47.3\\
GAL-fWSD512\cite{gal-fwsd}&58.4&63.8&45.8&24.0&22.7&67.7&65.7&58.9&15.0&58.1&47.0&53.7&23.8&64.3&36.2&22.3&46.7&50.3&70.8&55.1&47.5\\
ZLDN\cite{zigzag}&55.4&68.5&50.1&16.8&20.8&62.7&66.8&56.5&2.1&57.8&47.5&40.1&69.7&68.2&21.6&27.2&53.4&56.1&52.5&58.2&47.6\\
TS2C+FRCNN\cite{ts2c}&-&-&-&-&-&-&-&-&-&-&-&-&-&-&-&-&-&-&-&-&48.0\\
PCL-Ens.+FRCNN\cite{pcl}&63.2&69.9&47.9&22.6&27.3&71.0&69.1&49.6&12.0&60.1&51.5&37.3&63.3&63.9&15.8&23.6&48.8&55.3&61.2&62.1&48.8\\
ML-LocNet-L+\cite{ml-locanet}&60.8&70.6&47.8&30.2&24.8&64.9&68.4&57.9&11.0&51.3&55.5&48.1&68.7&69.5&28.3&25.2&51.3&56.5&60.0&43.1&49.7\\
WSRPN-Ens.+FRCNN\cite{wsrpn}&63.0&69.7&40.8&11.6&27.7&70.5&74.1&58.5&10.0&66.7&60.6&34.7&75.7&70.3&25.7&26.5&55.4&56.4&55.5&54.9&50.4\\
Multi-Evidence\cite{multi-evidence}&64.3&68.0&56.2&36.4&23.1&68.5&67.2&64.9&7.1&54.1&47.0&57.0&69.3&65.4&20.8&23.2&50.7&59.6&65.2&57.0&51.2\\
W2F+RPN+FSD2\cite{w2f}&63.5&70.1&50.5&31.9&14.4&72.0&67.8&73.7&23.3&53.4&49.4&65.9&57.2&67.2&27.6&23.8&51.8&58.7&64.0&62.3&52.4\\\hline
Ours-Ens.&59.8&72.8&54.4&35.6&30.2&74.4&70.6&74.5&27.7&68.0&51.7&46.3&63.7&68.6&14.8&27.8&54.9&60.9&65.1&67.4&54.5\\
\hline
\end{tabular}
\caption{Comparison of AP performance (\%) on PASCAL VOC 2007 test. The upper part shows results by \textbf{single end-to-end model}. The lower part shows results by \textbf{multi-phase approaches or ensemble model}.}
\label{table:soa_2007_map}
\end{table*}
\begin{table*}[]
\centering
\fontsize{7}{8}\selectfont
\def\arraystretch{1.1}
\setlength{\tabcolsep}{2.5pt}
\begin{tabular}{l|cccccccccccccccccccc|c}
\specialrule{.2em}{.1em}{.1em}
Methods&aero&bike&bird&boat&bottle&bus&car&cat&chair&cow&table&dog&horse&mbike&person&plant&sheep&sofa&train&tv&mAP
\\\hline
ContextLocNet\cite{contextlocnet}&64.0&54.9&36.4&8.1&12.6&53.1&40.5&28.4&6.6&35.3&34.4&49.1&42.6&62.4&19.8&15.2&27.0&33.1&33.0&50.0&35.3\\
OICR\cite{oicr}&67.7&61.2&41.5&25.6&22.2&54.6&49.7&25.4&19.9&47.0&18.1&26.0&38.9&67.7&2.0&22.6&41.1&34.3&37.9&55.3&37.9\\
Self-taught\cite{selftaught}&60.8&54.2&34.1&14.9&13.1&54.3&53.4&58.6&3.7&53.1&8.3&43.4&49.8&69.2&4.1&17.5&43.8&25.6&55.0&50.1&38.3\\
WCCN\cite{wccn}&-&-&-&-&-&-&-&-&-&-&-&-&-&-&-&-&-&-&-&-&37.9\\
TS2C\cite{ts2c}&67.4&57.0&37.7&23.7&15.2&56.9&49.1&64.8&15.1&39.4&19.3&48.4&44.5&67.2&2.1&23.3&35.1&40.2&46.6&45.8&40.0\\
WSRPN\cite{wsrpn}&-&-&-&-&-&-&-&-&-&-&-&-&-&-&-&-&-&-&-&-&40.8\\
PCL\cite{pcl}&58.2&66.0&41.8&24.8&27.2&55.7&55.2&28.5&16.6&51.0&17.5&28.6&49.7&70.5&7.1&25.7&47.5&36.6&44.1&59.2&40.6\\\hline
MIL-OICR+GAM+REG(Ours)&64.7&66.3&46.8&28.5&28.4&59.8&58.6&70.9&13.8&55.0&15.7&60.5&63.9&69.2&8.7&23.8&44.7&52.7&41.5&62.6&46.8\\
MIL-PCL+GAM+REG(Ours)&60.4&68.6&51.4&22.0&25.9&49.4&58.4&62.1&14.5&58.8&24.6&60.4&64.3&70.3&9.4&26.0&47.7&45.5&36.7&55.8&45.6\\\hline\hline
MELM\cite{multi-evidence}&-&-&-&-&-&-&-&-&-&-&-&-&-&-&-&-&-&-&-&-&42.4\\
OICR-Ens.+FRCNN\cite{oicr}&-&-&-&-&-&-&-&-&-&-&-&-&-&-&-&-&-&-&-&-&42.5\\
ZLDN\cite{zigzag}&54.3&63.7&43.1&16.9&21.5&57.8&60.4&50.9&1.2&51.5&44.4&36.6&63.6&59.3&12.8&25.6&47.8&47.2&48.9&50.6&42.9\\
GAL-fWSD512\cite{gal-fwsd}&64.9&56.8&47.0&18.1&22.2&60.0&51.7&60.7&12.9&43.1&23.6&58.5&52.1&66.9&39.5&19.0&39.6&36.1&62.7&27.4&43.1\\
ML-LocNet-L+\cite{ml-locanet}&53.9&60.4&40.4&23.3&18.7&58.7&63.3&52.5&13.3&49.1&46.8&33.5&61.0&65.8&21.3&22.9&46.8&48.1&52.6&40.4&43.6\\
TS2C+FRCNN\cite{ts2c}&-&-&-&-&-&-&-	&-&-&-&-&-&-&-&-&-&-&-&-&-&44.0\\
PCL-Ens.+FRCNN\cite{pcl}&69.0&71.3&56.1&30.3&27.3&55.2&57.6&30.1&8.6&56.6&18.4&43.9&64.6&71.8&7.5&23.0&46.0&44.1&42.6&58.8&44.2\\
WSRPN-Ens.+FRCNN\cite{wsrpn}&-&-&-&-&-&-&-&-&-&-&-&-&-&-&-&-&-&-&-&-&45.7\\
W2F+RPN+FSD2\cite{w2f}&73.0&69.4&45.8&30.0&28.7&58.8&58.6&56.7&20.5&58.9&10.0&69.5&67.0&73.4&7.4&24.6&48.2&46.8&50.7&58.0&47.8\\\hline
Ours-Ens.&66.8&71.1&56.0&28.4&34.2&56.2&60.3&63.8&17.3&61.3&24.8&59.7&67.4&73.6&12.0&30.0&52.7&47.1&45.9&61.5&49.5\\
\hline
\end{tabular}
\caption{Comparison of AP performance (\%) on PASCAL VOC 2012 test. The upper part shows results by \textbf{single end-to-end model}. The lower part shows results by \textbf{multi-phase approaches or ensemble model}.}
\label{table:soa_2012_map}
\end{table*}
\begin{table*}[]
\centering
\fontsize{7}{8}\selectfont
\def\arraystretch{1.1}
\setlength{\tabcolsep}{2.5pt}
\begin{tabular}{l|cccccccccccccccccccc|c}
\specialrule{.2em}{.1em}{.1em}
Methods&aero&bike&bird&boat&bottle&bus&car&cat&chair&cow&table&dog&horse&mbike&person&plant&sheep&sofa&train&tv&mAP\\\hline
WSDDN\cite{wsddn}&65.1&58.8&58.5&33.1&39.8&68.3&60.2&59.6&34.8&64.5&30.5&43.0&56.8&82.4&25.5&41.6&61.5&55.9&65.9&63.7&53.5\\
ContextLocNet\cite{contextlocnet}&83.3&68.6&54.7&23.4&18.3&73.6&74.1&54.1&8.6&65.1&47.1&59.5&67.0&83.5&35.3&39.9&67.0&49.7&63.5&65.2&55.1\\
OICR\cite{oicr}&81.7&80.4&48.7&49.5&32.8&81.7&85.4&40.1&40.6&79.5&35.7&33.7&60.5&88.8&21.8&57.9&76.3&59.9&75.3&81.4&60.6\\
Self-taught\cite{selftaught}&72.7&55.3&53.0&27.8&35.2&68.6&81.9&60.7&11.6&71.6&29.7&54.3&64.3&88.2&22.2&53.7&72.2&52.6&68.9&75.5&56.1\\
WCCN\cite{wccn}&83.9&72.8&64.5&44.1&40.1&65.7&82.5&58.9&33.7&72.5&25.6&53.7&67.4&77.4&26.8&49.1&68.1&27.9&64.5&55.7&56.7\\
TS2C\cite{ts2c}&84.2&74.1&61.3&52.1&32.1&76.7&82.9&66.6&42.3&70.6&39.5&57.0&61.2&88.4&9.3&54.6&72.2&60.0&65.0&70.3&61.0\\
WSRPN\cite{wsrpn}&77.5&81.2&55.3&19.7&44.3&80.2&86.6&69.5&10.1&87.7&68.4&52.1&84.4&91.6&57.4&63.4&77.3&58.1&57.0&53.8&63.8\\
PCL\cite{pcl}&79.6&85.5&62.2&47.9&37.0&83.8&83.4&43.0&38.3&80.1&50.6&30.9&57.8&90.8&27.0&58.2&75.3&68.5&75.7&78.9&62.7\\\hline
MIL-OICR+GAM+REG(Ours)&81.7&81.2&58.9&54.3&37.8&83.2&86.2&77.0&42.1&83.6&51.3&44.9&78.2&90.8&20.5&56.8&74.2&66.1&81.0&86.0&66.8\\
MIL-PCL+GAM+REG(Ours)&80.0&83.9&74.2&53.2&48.5&82.7&86.2&69.5&39.3&82.9&53.6&61.4&72.4&91.2&22.4&57.5&83.5&64.8&75.7&77.1&68.0\\\hline
\hline
PDA \cite{pda} & 78.2 & 67.1 & 61.8 & 38.1 & 36.1 & 61.8 & 78.8&55.2& 28.5 & 68.8 & 18.5 & 49.2 & 64.1 & 73.5 & 21.4 & 47.4 & 64.6&22.3 &60.9 & 52.3 & 52.4\\
WSDDN-Ens. \cite{wsddn} &68.9 & 68.7 & 65.2 & 42.5 & 40.6 & 72.6 & 75.2 & 53.7 &29.7 & 68.1 &33.5 & 45.6 & 65.9 & 86.1 & 27.5 & 44.9 & 76.0 & 62.4 & 66.3 & 66.8 &58.0 \\
OICR-Ens.+FRCNN \cite{oicr} &85.8 & 82.7 & 62.8 & 45.2 &43.5 & 84.8 &87.0 & 46.8 & 15.7 & 82.2 & 51.0 & 45.6 & 83.7 & 91.2 &22.2 & 59.7 &75.3 & 65.1 & 76.8 & 78.1 & 64.3  \\GAL-fWSD \cite{gal-fwsd}& - & - & - & - & - & - & -	& - & -& - & -& -& - & - & - & - & - & - & - & 	- &  67.2 \\
ZLDN \cite{zigzag} & 80.3 & 76.5 & 64.2 & 40.9 & 46.7 &78.0 & 84.3 &57.6 & 21.1 & 69.5 & 28.0 & 46.8 & 70.7 & 89.4 & 41.9 &54.7 & 76.3 &61.1 & 76.3 & 65.2 & 61.5\\
PCL-Ens.+FRCNN \cite{pcl} &83.8 & 85.1 & 65.5 & 43.1 & 50.8& 83.2 &85.3 & 59.3 & 28.5 & 82.2 & 57.4 & 50.7 & 85.0 & 92.0 &27.9 & 54.2 &72.2 & 65.9 & 77.6 & 82.1 & 66.6\\
ML-LocNet-L+\cite{ml-locanet} & 88.1 & 85.5 & 71.2 & 49.4 &57.4 & 90.7& 77.6 & 53.5 & 42.6 & 79.6 & 34.1 & 69.1 & 81.7 & 91.9 &35.4 & 64.6 &79.3 & 64.3 & 79.3 & 69.6 & 68.2\\
WSRPN-Ens.+FRCNN \cite{wsrpn} &83.8 & 82.7 & 60.7 & 35.1 &53.8 & 82.7& 88.6 & 67.4 & 22.0 & 86.3 & 68.8 & 50.9 & 90.8 & 93.6 &44.0 & 61.2 &82.5 & 65.9 & 71.1 & 76.7 & 68.4\\
W2F+RPN+FSD2 \cite{w2f} & 85.4 & 87.5 & 62.5 & 54.3 & 35.5& 85.3 &86.6 & 82.3 & 39.7 & 82.9 & 49.4 & 76.5 & 74.8 & 90.0 &46.8 & 53.9 &84.5 & 68.3 & 79.1 & 79.9 & 70.3\\\hline
Ours-Ens.  & 83.3  & 85.5  & 68.8  & 56.9  & 49.6  & 84.3  &87.0  & 83.1  & 44.2  & 86.3  & 55.5  & 54.4  & 81.6  & 92.8  & 22.8  &60.4  & 81.4  & 70.2  & 81.4  & 81.4  & \textbf{70.6} \\
\hline
\end{tabular}
\caption{Comparison of correct localization (CorLoc) (\%) on PASCAL VOC 2007 trainval. The upper part shows results by \textbf{single end-to-end model}. The lower part shows results by \textbf{multi-phase approaches or ensemble model}.}
\label{table:soa_2007_corloc}
\end{table*}
\begin{table*}[]
\centering
\fontsize{7}{8}\selectfont
\def\arraystretch{1.1}
\setlength{\tabcolsep}{2.5pt}
\begin{tabular}{l|cccccccccccccccccccc|c}
\specialrule{.2em}{.1em}{.1em}
Methods&aero&bike&bird&boat&bottle&bus&car&cat&chair&cow&table&dog&horse&mbike&person&plant&sheep&sofa&train&tv&mAP\\\hline
ContextLocNet\cite{contextlocnet}&78.3&70.8&52.5&34.7&36.6&80.0&58.7&38.6&27.7&71.2&32.3&48.7&76.2&77.4&16.0&48.4&69.9&47.5&66.9&62.9&54.8\\
OICR\cite{oicr}&86.2&84.2&68.7&55.4&46.5&82.8&74.9&32.2&46.7&82.8&42.9&41.0&68.1&89.6&9.2&53.9&81.0&52.9&59.5&83.2&62.1\\
Self-taught\cite{selftaught}&82.4&68.1&54.5&38.9&35.9&84.7&73.1&4.8&17.1&78.3&22.5&57.0&70.8&86.6&18.7&49.7&80.7&45.3&70.1&77.3&58.8\\
TS2C\cite{ts2c}&79.1&83.9&64.6&50.6&37.8&87.4&74.0&74.1&40.4&80.6&42.6&53.6&66.5&88.8&18.8&54.9&80.4&60.4&70.7&79.3&64.4\\
WSRPN\cite{wsrpn}&-&-&-&-&-&-&-&-&-&-&-&-&-&-&-&-&-&-&-&-&64.9\\
PCL\cite{pcl}&77.2&83.0&62.1&55.0&49.3&83.0&75.8&37.7&43:2&81.6&46:8&42.9&73.3&90.3&21.4&56.7&84.4&55.0&62.9&82.5&63.2\\\hline
MIL-OICR+GAM+REG(Ours)&82.4&83.7&72.4&57.9&52.9&86.5&78.2&78.6&40.1&86.4&37.9&67.9&87.6&90.5&25.6&53.9&85.0&71.9&66.2&84.7&69.5\\
MIL-PCL+GAM+REG(Ours)&80.2&83.0&73.1&51.6&48.3&79.8&76.6&70.3&44.1&87.7&50.9&70.3&84.7&92.4&28.5&59.3&83.4&64.6&63.8&81.2&68.7\\
\hline\hline
OICR-Ens.+FRCNN \cite{oicr}  & - & - & - & - & - & - & - & - &- & - & - & - & - & - & - & - & - & - & - & - & 65.6  \\
ZLDN \cite{zigzag} & 80.3 & 76.5 & 64.2 & 40.9 & 46.7 & 78.0 & 84.3 &57.6 & 21.1 & 69.5 & 28.0 & 46.8 & 70.7 & 89.4 & 41.9 & 54.7 & 76.3 &61.1 & 76.3 & 65.2 & 61.5\\
GAL-fWSD512 \cite{gal-fwsd}& - & - & - & - & - & - & - & - &- & - & - & - & - & - & - & - & - & - & - & - & 67.2 \\
ML-LocNet-L+\cite{ml-locanet} & 88.1 & 85.5 & 71.2 & 49.4 & 57.4 & 90.7& 77.6 & 53.5 & 42.6 & 79.6 & 34.1 & 69.1 & 81.7 & 91.9 & 35.4 & 64.6 &79.3 & 64.3 & 79.3 & 69.6 & 68.2\\
PCL-Ens.+FRCNN \cite{pcl} & 86.7 &86.7 &74.8 &56.8 & 53.8 & 84.2 & 80.1& 42.0 & 36.4 & 86.7 & 46.5 & 54.1 & 87.0 & 92.7 & 24.6 & 62.0 & 86.2 &63.2 & 70.9 &84.2 & 68.0 \\
WSRPN-Ens.+FRCNN \cite{wsrpn} & - & - & - & - & - & - & - & - &- & - & - & - & - & - & - & - & - & - & - & - & 69.3\\
W2F+RPN+FSD2 \cite{w2f} & 88.8 & 85.8 & 64.9 & 56.0 & 54.3 & 88.1 &79.1 & 67.8 & 46.5 & 86.1 & 26.7 & 77.7 & 87.2 & 89.7 & 28.5 & 56.9 &85.6 & 63.7 & 71.3 & 83.0 & 69.4\\\hline
Ours-Ens.  & 82.0  & 85.1  & 73.7  & 56.6  & 53.0  & 85.8  &79.2  & 80.9  & 46.0  & 87.7  & 46.2  & 72.7  & 88.2  & 91.6  & 26.0  &60.6  & 83.7  & 72.2  & 67.8  & 85.0  & \textbf{71.2}  \\ \hline
\end{tabular}
\caption{Comparison of correct localization (CorLoc) (\%) on PASCAL VOC 2012 trainval. The upper part shows results by \textbf{single end-to-end model}. The lower part shows results by \textbf{multi-phase approaches or ensemble model}.}
\label{table:soa_2012_corloc}
\end{table*}
\subsection{Multi-Task Branch}
After pseudo GT annotations are generated, a multi-task branch can operate
fully supervised classification and regression as Fast R-CNN \cite{fast-rcnn}. The detection branch has two sibling branches. The first branch predicts a discrete probability distribution (per RoI), $p\in \mathbb{R}^{(C+1) \times 1}$, over $C$+$1$ categories, which is computed by a softmax over the  $C$+$1$ outputs of a FC layer. The second sibling branch outputs bounding-box regression offsets, $t^c = (t^c_x, t^c_y, t^c_w, t^c_h )$ for each of the $C$ object classes, indexed by $c$.

Since we get the instance annotations from MIL branch as introduced in
Section \ref{sec:pgm}, each RoI now has a GT bounding-box regression target $v$ and GT classification target $u$. We use a multi-task loss $\mathcal{L}_{det}$ of all labeled RoIs for classification and bounding-box regression:
\begin{equation}\label{eq:loss_det}
\mathcal{L}_{det} = \mathcal{L}_{cls}  +  \lambda\mathcal{L}_{loc},
\end{equation}
\noindent where $\mathcal{L}_{cls}$ is classification loss, and $\mathcal{L}_{loc}$ is regression loss. $\lambda$ controls the balance between two losses. For $\mathcal{L}_{loc}$, smooth $L_1$ loss is used. For $\mathcal{L}_{cls}$, since the pseudo GT annotations are noisy, we add a weight $w^r$ with respect to RoI $r$:
\begin{equation}\label{eq:loss_det}
\mathcal{L}_{cls} =
- \frac{1}{|R|}\sum_{r=1}^{|R|}\sum_{c=1}^{C+1} w^r u_c^r \log p_c^r,
\end{equation}
\noindent where $|R|$ is the number of proposals. The weight $w^r$ is calculated following the weights calculation method in \cite{oicr} when refining the classifiers.

The overall network is trained by optimizing the following composite loss
functions from the four components using stochastic gradient descent:
\begin{equation}\label{eq:loss_det}
\mathcal{L} = \mathcal{L}_{img_{cls}} + \mathcal{L}_{mil} + \mathcal{L}_{refine} +
\mathcal{L}_{det},
\end{equation}
where $\mathcal{L}_{img_{cls}}$ is the multi-label classification
loss of GAM; $\mathcal{L}_{mil}$ is the multi-label classification loss of WSDDN; $\mathcal{L}_{refine}$ is the classifier refinement loss; and $\mathcal{L}_{det}$ is multi-task loss of the detection sub-network.
\section{Experiments}
In this section, we first introduce the evaluation datasets and the implementation details of our approach. Then we explore the contributions of each proposed module by the ablation experiments. Finally, we compare the performance of our method with the-state-of-the-art methods.
\subsection{Datasets and Evaluation Metrics}
We evaluate our method on the popular PASCAL VOC 2007 and 2012 datasets \cite{voc} which have 9963 and 22531 images  for 20 object classes, respectively. These two datasets are split into train, validation, and test sets. We use the trainval set (5011 images for 2007 and 11540 for 2012) for training. As we focus on weakly supervised detection, only image-level labels are utilized during training. Average Precision (AP) and the mean of AP (mAP) are taken as the evaluation metrics to test our model on the testing set. Correct localization (CorLoc) is also used to evaluate our model on the trainval set to measure the localization accuracy \cite{wsddn}. Both metrics are evaluated on the PASCAL criteria, i.e., IoU $>$ 0.5 between ground truths boxes and predicted boxes.
\begin{figure*}[ht]
	\centering
	\includegraphics[width = 1.0\textwidth]{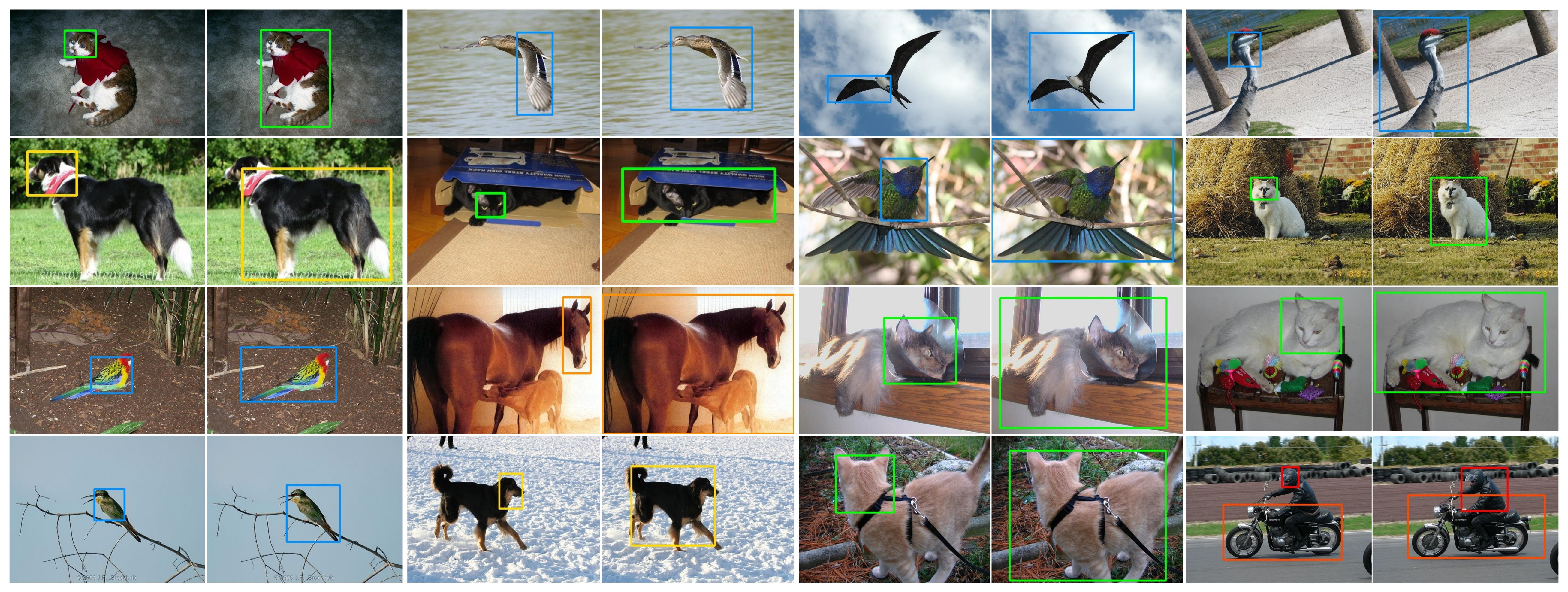}
	\caption{Qualitative detection results of our method and the baseline (OICR+FRCN).The results of baseline are shown in the odd columns. The results of our method are shown in even columns.}
	\label{fig:vis}
\end{figure*}
\subsection{Implementation Details}
We use the object proposals generated by selective search windows \cite{ss} and adopt VGG16 \cite{vgg} pre-trained on ImageNet \cite{imagenet} as the backbone of our proposed network.

For the newly added layers, the parameters are randomly initialized with a Gaussian distribution $\mathcal{N}(\mu,\delta)(\mu=0,\delta=0.01)$ and 10 times learning rate. During training, we adopt a mini-batch size of 2 images, and set the learning rate to 0.001 for the first 40K iterations and then decrease it to 0.0001 in the following 30K iterations. The momentum and weight decay are set to 0.9 and 0.0005, respectively. We use five image scales , i.e., $\{480, 576, 688, 864, 1200\}$, and horizontal flips for both training and testing data augmentation. During testing, we use the mean output of the regression branch, including classificaiton scores and bounding boxes, as the final results. Our experiments are based on the deep learning framework of Caffe \cite{caffe}. All of the experiments run on NVIDIA GTX 1080Ti GPUs.
\subsection{Ablation Studies}
We conduct ablation experiments on PASCAL VOC 2007 to prove the effectiveness of our proposed network. We validate the contribution of each component including GAM and regression branch.
\subsubsection{Baseline}
The baseline is the \textbf{MIL} detector without GAM and regression branch that we introduced in Section \ref{sec:gam}, which is the same as \textbf{OICR} \cite{oicr}. We re-run the experiment and get a slightly higher result of 41.3\% mAP (41.2\% mAP in \cite{oicr}).
\subsubsection{Guided Attention Module}
To verify the effect of GAM, we conduct experiments with and w/o GAM. We denote the network with GAM  as \textbf{MIL+GAM}, which does not include regression branch. From Table \ref{table:ablation_study}, we can conclude that GAM does help the detector learn better features and improves the accuracy of MIL detector by 2.0\%.
\subsubsection{Joint Optimization}
\begin{figure*}[ht]
	\centering
	\includegraphics[width = 1\textwidth]{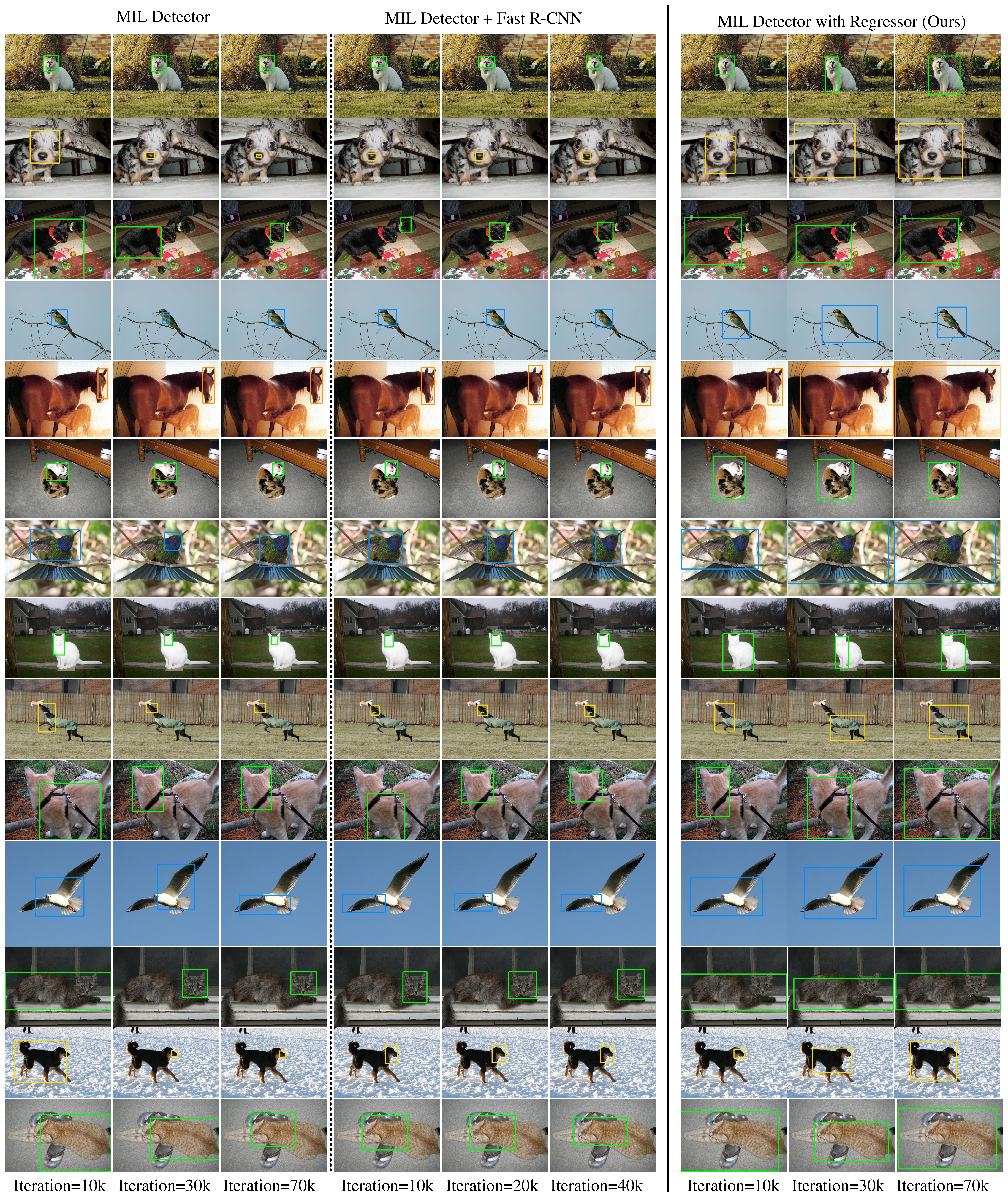}
	\caption{Detection results of MIL detector (left part), Fast R-CNN with
		pseudo GT from MIL detector (middle part) and our jointly training
		network (right part)  at different training
		iterations .}
	\label{fig:vis_supp}
\end{figure*}

To optimize proposal classification and regression jointly, we propose to use bounding-box regression in an online manner together with MIL detection. To verify the effect of online regression, we conduct control experiments under two setting: 1) our joint optimization of MIL detector and regressor, which we denote as \textbf{MIL+REG}; 2) we train a MIL detector first, then use the pseudo GT from the MIL detector to train a fully supervised Fast R-CNN \cite{fast-rcnn}.
We denote this setting as \textbf{MIL+FRCN}. The experimental results are summarized in Table \ref{table:ablation_study}. From the results, we can see the performance of our \textbf{MIL+REG} is much higher than \textbf{MIL+FRCN}. We attribute the improvements to joint optimization. Separate optimization of MIL detector and regressor
result in sub-optimal results. It easily gets stuck in local minima if the pseudo GTs are not accurate. This can be seen from the results of the object category \textit{cat} and \textit{dog}.  The two object classes are much easier to over-fit to the discriminate parts in the MIL detection. Our joint optimization strategy can alleviate this problem as shown in Figure \ref{fig2}. More visualization results are shown in the supplementary file.
We also carry the exploration study on the CorLoc metric, as reported in Table \ref{table:ablation_study_corloc}. From these results, we can draw the same conclusion. In Figure \ref{fig:vis_supp}, we show more qualitative results in the same way to supplement Figure \ref{fig2}.
\subsection{Comparison with State-of-the-Art}
To fully compare with other methods, we report the results for both ``\textbf{single end-to-end network}'' and ``\textbf{multi-phase approaches or ensemble model}''. The results on VOC 2007 and VOC 2012 are shown in Table \ref{table:soa_2007_map}, Table \ref{table:soa_2007_corloc}, Table \ref{table:soa_2012_map} and Table \ref{table:soa_2012_corloc}. From the tables, we can see that our method achieves the highest performance, outperforming the state-of-the-arts for both cases. \textbf{It is worth noting that our single model results are even much better than the ensemble models results of most methods which ensemble the results of multiple CNN networks.} For example, compared with OICR \cite{oicr}, which we use as baseline, \textbf{our single model outperforms the ensemble models of OICR significantly while keeping much lower complexity} (47.0\% mAP Versus 48.6\% mAP; 60.6\% CorLoc Versus 66.8\% CorLoc on VOC 2007). In Figure \ref{fig:vis}, we also illustrate some detection results by our network as compared to those by our baseline method, i.e., OICR+FRCN. It can be concluded from the illustration that our joint training strategy significantly alleviates the detector focusing on the most discriminative parts.
\subsection{Discussion}
C-WSL \cite{cwsl} also explored bounding box regression in weakly supervised object detection network. We list the relationship and some differences below. \emph{Relationship}: We both use bounding box regression in an online manner. However, there are key differences in network architecture between the two, which lead to the performance of C-WSL being much lower than ours, even though they use additional object count labels. \emph{Differences:} The network structure is different. We use bounding box regression after several box classifier refinements and use only once. C-WSL \cite{cwsl} uses a box regressor together with each box classifier refinement after the MIL branch. \emph{Their structure brings two problems.} First, a single MIL branch's classification performance is very poor, it is not wise to directly use the box regressor to refine the box location after the MIL branch. The second problem is that the bounding box regression is used in a cascade manner for each refinement without re-extracting features for the RoIs. Specifically, the subsequent box regression branch should take the refined box locations from the previous box regression branch to update RoIs and re-extracting RoIs features for the classifier and regressor. Because of the above problems, after deducting the improvement of extra label information, their network only improves 1.5\% compared with OICR as shown in \cite{cwsl} while our network has increased by 6\% compared with OICR (Please note that we use the same set of code released by the authors of OICR). In addition, \cite{cwsl} does not solve the problem of local minima. On the two categories that most affected by the local minima problem, \cite{cwsl} drops 4\% in the dog category and improves 3\% in the cat category while our method improves 16.3\% and 38.6\% respectively.\\
\section{Conclusion}
In this paper, we present a novel framework for weakly supervised object detection. Different from traditional approaches in this field, our method jointly optimize the MIL detection and regression in an end-to-end manner. Meanwhile, a guided attention module is also added for better feature learning. Experiments show substantial and consistent improvements by our method. Our learning algorithm is potential to be applied in many other weakly supervised visual learning tasks.
\section*{Acknowledgements}
This work is supported by the National Key Research and Development Program of China under Grant No.2018YFB2101100 and the National Natural Science Foundation of China under Grants 61732018, U1435219  and 61802419.
{\small
\bibliographystyle{ieee_fullname}
\bibliography{ref}
}
\end{document}